# Towards Automatic Sizing for PPE with a Point Cloud Based Variational Autoencoder

Jacob, A. Searcy*

University of Oregon jsearcy@uoregon.edu

Susan L. Sokolowski

University of Oregon, ssokolow@uoregon.edu

**ABSTRACT**

Sizing and fitting of Personal Protective Equipment PPE is a critical part of the product creation process; however, traditional methods to do this type of work can be labor intensive and based on limited or non-representative anthropomorphic data. In the case of PPE, a poor fit can jeopardize an individual's health and safety. In this paper we present an unsupervised machine learning algorithm that can identify a representative set of exemplars, individuals that can be utilized by designers as idealized sizing models. The algorithm is based around a Variational Autoencoder (VAE) with a Point-Net inspired encoder and decoder architecture trained on Human point-cloud data obtained from the CEASAR dataset. The learned latent space is then clustered to identify a specified number of sizing groups. We demonstrate this technique on scans of human faces to provide designers of masks and facial coverings a reference set of individuals to test existing mask styles.

## 1. Introduction

Personnel Protective Equipment (PPE) is worn to mitigate exposure to hazards that can cause illness and injury. Poorly sized PPE can decrease effectiveness and jeopardize the safety of anyone who may not fit into existing systems. During the COVID-19 pandemic medical grade N95 masks and face coverings have become critical for first responders and the population at large, where the supply of properly fitting masks have impacted the pandemic response. N95 mask fittings can also be lengthy and require certified personnel. Demographic groups such as Woman and Asian health care workers have had difficulty finding good fit with existing mask designs, which places these workers at greater risk [8]. While COVID-19 has highlighted concerns around protective mask fit, concerns apply to all PPE equipment, and it is vital that we re-examine the sizing and development processes used to manufacture these products to ensure equitable safety to everyone.

In this paper we refer to 'sizing' to describe the process a user will undergo to select the best fitting classification of PPE. We use 'fit' to refer how the PPE will interface with the body and other worn PPE, and use 'classification' of PPE to refer to a specific product type for example small, medium, or large sizes. Sizing PPE requires selecting a product that must fit any number of complex bodily surfaces that are often not easily expressed with simple linear measurements or size classes. In this paper we will utilize Machine Learning (ML) as a new way of creating classifications for PPE that will allow designers to create new products and eventually allow us to automated sizing process.

We utilized an unsupervised ML algorithm and a databases of 3D scans to understand the variety of human features and identify exemplar scans that can serve as design targets for new classifications of PPE. Ultimately manufacturers will be able to create products for these classifications, and our same algorithms can be used with other personal scans to size products safely with better fits. This paper presents our ML model for determining design exemplars for face masks. In Section 2 we will describe the training data used by this project. In section 3 we will describe the variational autoencoder (VAE) based model utilized by this project. Section 4 describes the selection and evaluation of exemplars for identifying classifications, and Section 5 provides a summary and description of future work.

## 2. Processing of CAESAR Dataset

This study utilizes the Civilian American and European Surface Anthropometry Resource CAESAR dataset [2]. This dataset available through Shape Analysis Ltd. contains thousands of body scans from the United States along with associated demographic data and hand annotated anthropomorphic landmarks. We extract (X, Y, Z) point cloud coordinates for each scan from the vertices stored in each associated PLY file. In order to investigate mask fitting we isolate points belonging to the face which limits extraneous information seen by our ML model. In order to isolate the face of each participant we utilize three landmark points provided with the CEASAR data: the cervicale, a point located at the base of the neck, the left Tragion, and the right Tragion (points located on or near the ear). To reduce variation in head positioning we rotate each scan so the vector between left and right tragion is aligned with the y-axis. We then utilize only points that are forward of a vertical plane defined by the tragion and the floor and are also within 4 cm or above the cervicale. We found that it was necessary to allow some points below the cervicale to avoid truncating the chin of some participants. The selected points are then centered to remove variations in height between participants. Finally, these point clouds are resampled by randomly selecting 10,000 points to create a fixed size dataset. Figure 1 shows an example of the points extracted from an original CAESAR scan.

---

* Corresponding Author

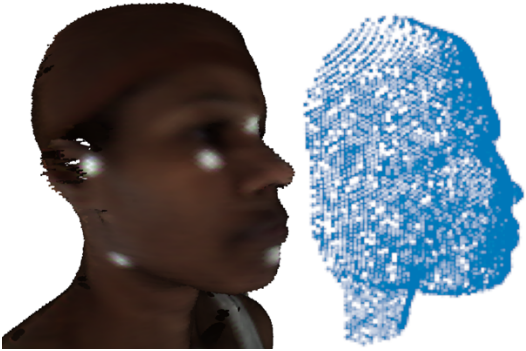

**Figure 1:** Example CAESAR Scan (left) with landmarks denoted by white dots, and extracted point could (left).

# 3. Variational Autoencoder

Variational Autoencoders (VAE)[6,9] are non-linear dimensionality reduction techniques, that utilizing deep neural networks. Our VAE utilizes a series of convolutional and fully-connect layers to encode an example point cloud into a latent space with reduced dimension, subsequently a decoder utilizes the latent space features to reconstruct the original point cloud. The model is trained on two objects. 1) a reconstruction objective that minimizes the difference between the original and reconstructed point cloud, and 2) a latent space objective that tends the learned latent space to be uniformly distributed.

The goal of this project is to provide understandable and useful information to product manufacturers, so of critical importance is the size of the latent space. In order to produce visualizations that can be quickly examined we choose a small value of three dimensions. While we found that larger latent spaces provided better point-cloud reconstructions, understanding and clustering in the higher dimensional proved more difficult.

## 3.1 Encoder/Decoder Network

We utilize an encoder and decoder architecture inspired by point-NET[3], and VAE's investigated by Point-NET author C. Qi[7]. The architecture utilized by our model is listed in Table 1. Our encoder utilizes a series of convolutions with a kernel size of one and can be thought of as template points, since each convolution acts on only one point at a time. This is a requirement for our encoder because the utilized point cloud data has no inherent ordering, and relationships between neighboring points in our input sequences are arbitrary. Another permutation invariant function a global max pooling layer is used to reduce the convolutions to a fixed length set of features, which are used to derive the latent space variables. Our decoder network utilizes similar fully connected networks, up-sampling layers that simply repeat existing elements a fixed number of times, and 1-dimensional convolutions. Un-like our encoder network we allow the decoder to learn a preferred point ordering and utilize neighboring sequence points during its generation process.

**Table 1. Module Layer Architecture with layer size representing the number of neurons for fully collected layers or the number filters x kernel size for convolutional layers.**

| Layer | Layer Size | Output Size |
|---|---|---|
| Encoder Network | | |
| Input | | 10,000 x 3 |
| 1D Conv+ PReLU | 1024 x 1 | 10,000x1024 |
| 1D Conv+ PReLU | 1024 x 1 | 10,000x1024 |
| 1D Conv+ PReLU | 1024 x 1 | 10,000x1024 |
| Global Max Pool | | 1,024 |
| Dense Latent Space | 3 | 3 |
| Decoder Network | | |
| Dense + PReLU | 1000 | 1,000 |
| Dense + PReLU | 1000 | 1,000 |
| Dense + PReLU | 100 | 3,000 |
| Reshape | | 1000x3 |
| Up sampling 2x | | 2,000x3 |
| 1D Conv. + PReLU | 1024x10 | 2,000x1024 |
| Up sampling 5x | | 10,000x1024 |
| 1D Conv. + PReLU | 200x50 | 10,000x200 |
| 1D Conv. + PReLU | 200x50 | 10,000x200 |
| 1D Conv. + PReLU | 200x50 | 10,000x200 |
| 1D Conv | 3x50 | 10,000x3 |

## 3.2 Model Objectives

Two loss objects are considered while training a VAE a reconstruction term and a latent space object. For our reconstruction term we utilize an approximation of the earth mover distance utilized by Fan et al. [4]. This loss, given by

$$L_r = \min_{\phi:S_1 \to S_2} \sum_{x \in S_1} \|x - \phi(x)\|_2$$

computes the squared distance between each point in the reconstructed point cloud $S_1$ and its counterpart point in the original point cloud $S_2$ given by the bijection $\phi(x): S_1 \to S_2$. During training we approximate the optimal bijection using a parallel method given by Bertsekas [1].

Several latent space objects exist that yield different representations we utilize the Maximum Mean Discrepancy (MMD) proposed for use in an info-VAE by Zhao *et al*. [10]. This objective is preferred over the more commonly used evidence lower bound (ELBO), because the ELBO has been show to over-estimate variance, and in some cases provide un-informative latent spaces. The MMD objective is robust these failings and provides good performance without having to add additional hyperparameters to balance the latent space objective and the reconstruction objective. The MMD loss utilizes 'the kernel trick' to compare the moments of the latent space distribution $p(z)$ to that of a target distribution $q(z)$.

$$L_l = \mathbb{E}_{p(z),p(z')}[k(z,z')] - 2\mathbb{E}_{q(z),p(z')}[k(z,z')] + \mathbb{E}_{q(z),q(z')}[k(z,z')]$$

This loss term becomes identically zero when $p(z) = q(z)$ we utilize a simple uniform distribution for $q(z)$, and a gaussian kernel.

$$k(z,z') = e^{-\frac{(z-z')^2}{2}}$$

During training we minimize the combined loss

$$L = L_l + L_r$$

utilizing ADAM [5] to preform stochastic gradient descent until the loss calculated on a withheld data sets stops improving. The model was train on a NIVIDA Titan RTX GPU over the course of a few days.

## 3.3 Learned Latent Space

After training we explored the relationships encoded by each latent dimension by comparing representative data points. We started by defining the dataset average face, by identifying the point cloud that has the smallest Euclidean distance to the mean of the latent space distribution. $f_{avg} = argmin_i \|z_i - \bar{z}\|_2$. Starting from the population mean we set the value of one dimension to the 5th percentile and then the 95th percentile two create two more latent space points for each dimension. We then found the nearest data point to each latent space point. These data points are plotted if Figure 2 with each point cloud highlighted by the distance of each point to the nearest point in $f_{avg}$. Dimension-one of the latent space relates closely to the width of the face dimension-two encodes the location of the chin and nose, and dimension-three relates closely to the overall size of the face. Also displayed are the reconstructed point clouds created from each of the corresponding latent space points. These point clouds lack many of the features encoded by the original 30,000 input features, but still reproduce the shape of each input scan. Since the face shape is of primary importance to good mask fit, we accept this latent space as one useful for size selection.

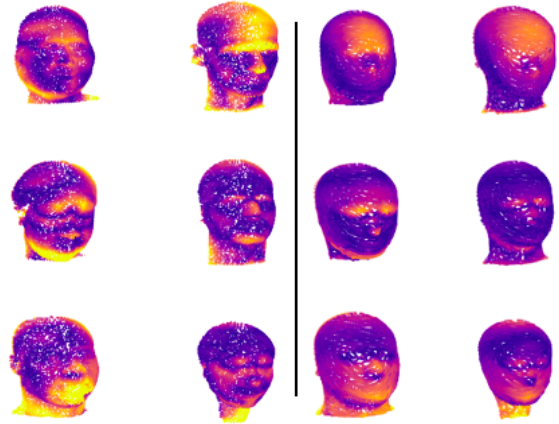

**Figure 2: Example point clouds representing the features encoded by each of the three latent dimensions. The left set of images are the original point clouds, while the right set are the reconstruction models from the three latent spaces. Within each set the left column are exemplars with a latent value at the 5th percentile for dimension 1 in row 1 dimension 2 in row 2 and dimension 3 in row 3. All other latent values are close to the population mean. Each point cloud is highlighted by the distance to the nearest point in the exemplar that sits closest to the global mean.**

## 4. Size Selections

While useful representations, the reconstructed scans are not real point clouds that can serve as sizing targets. This motivates using 'exemplars' which we define as the data point with the minimum Euclidean distance to a given latent space point. We wanted to select a set of exemplars that can broadly represent our dataset. In order select these exemplars, we utilized a simple k-means clustering in the latent space. This allowed us to specify several classification targets by specifying k. There is a tradeoff between the number of classifications selected and the work required for designing and manufacturing a product for each classification. For this example, we utilized a k=3. Each cluster center identified by the model can then be associated with an exemplar.

We aimed to create as complete a set of exemplars as possible for use by PPE designers. The CAESAR dataset contains demographic information on race with categories defined as African American, Asian, Spanish, and White and information on gender with categories Female and Male. White is the most prevalent race in the dataset followed by African American with equal splits between Male and Female. To avoid our clustering algorithm focusing primarily on white individuals we simply provide exemplars for each race and gender combinations. This provides and inclusive set of exemplars for designers to utilize during the design process. Figure 3 shows an exemplar clustering for the white female group. The full set of three exemplars for each of the eight demographics are currently being analyzed by designers to evaluate the quality of fit for several available mask designs.

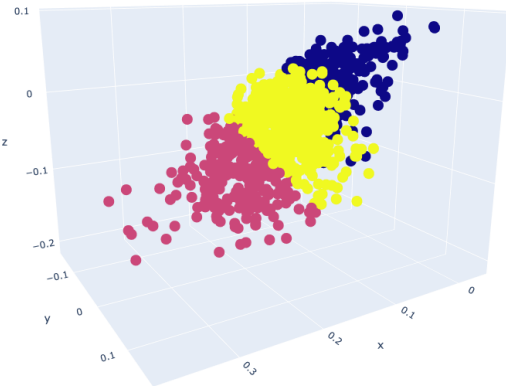

**Figure 3: Clustering Examples** Example of clustering in latent space with three size groups for the White and Female data points.

## 5. Summary and Future Work

We have presented a variational auto-encoder algorithm that can learn small and useful latent spaces for complicated 3D-point clouds. We applied this algorithm to provide a broad set of exemplars to use as size targets for product manufacturers. These targets are inclusive of race and gender, and account for the variation across a broad population. This work is enabling University of Oregon product innovators to evaluate existing mask designs and identify potential gaps in current sizing. We foresee this work expanding to not just providing exemplars to designers but allowing for 'virtual' sizing's where personal scans can be encoded by our model and return products designed specifically for that user's features. This has the potential to decrease time demands on trained staff and provide better fitting equipment for safer workplaces and activities.

### ACKNOWLEDGMENTS

The authors gratefully acknowledge the University of Oregon's Data Science Initiative Seed Fund for support of this work. In addition, this work benefited from access to the University of Oregon high performance computer, Talapas.